# Can Sentiment Analysis Reveal Structure in a "Plotless" Novel?

Katherine Elkins[1] and Jon Chun
Kenyon College

**Abstract**

Modernist novels are thought to break with traditional plot structure. In this paper, we test this theory by applying Sentiment Analysis to one of the most famous modernist novels, Virginia Woolf's *To the Lighthouse*. We first assess Sentiment Analysis in light of the critique that it cannot adequately account for literary language: we use a unique statistical comparison to demonstrate that even simple lexical approaches to Sentiment Analysis are surprisingly effective. We then use the Syuzhet.R package to explore similarities and differences across modeling methods. This comparative approach, when paired with literary close reading, can offer interpretive clues. To our knowledge, we are the first to undertake a hybrid model that fully leverages the strengths of both computational analysis and close reading. This hybrid model raises new questions for the literary critic, such as how to interpret relative versus absolute emotional valence and how to take into account subjective identification. Our finding is that while *To the Lighthouse* does not replicate a plot centered around a traditional hero, it does reveal an underlying emotional structure distributed between characters—what we term a **distributed heroine model**. This finding is innovative in the field of modernist and narrative studies and demonstrates that a hybrid method can yield significant discoveries.

**A Brief Introduction to Sentiment Analysis**

Today, Sentiment Analysis, with its unusually vibrant research community, is one of the most widely deployed Natural Language Processing (NLP) techniques.[2] Because of recent advances, Sentiment Analysis can now be used to monitor drivers in self-driving vehicles, advise and guide customer service conversations, and create emotionally intelligent chatbots like Xiaoice that humans find more engaging than other humans. Also known as Opinion Mining and AffectiveAI, Sentiment Analysis is now used in countless production and research systems to detect, model, predict and even shape future human preferences and actions.

---

[1] National Endowment for the Humanities Distinguished Teaching Professor, Digital Humanities
[2] Computational approaches to Sentiment Analysis have been employed at least since the 1960's (Stone and Anderson). In Computational Linguistics, Sentiment Analysis has been growing as a subfield for several decades (Kim and Klinger). Recently, statistical data-driven approaches (e.g. Word Embeddings, GPT-2) have proven far superior to traditional rule-based approaches found in formal grammars in the Chomsky Hierarchy.

Multimodal Sentiment Analysis, using advanced machine learning and deep neural network technologies (Barnes et al.) analyzes both written words and audio-visual content to ascertain emotional arc in film (Chu et al.). Sentiment Analysis is also a uniquely useful technology for uncovering aspects of latent structure and enabling a comparative study at the scale of the novel. Many NLP techniques like Parts of Speech and Name Entity Recognition operate on local syntax, while the most sophisticated State of the Art (SOTA) NLP generally relies on Deep Neural Network approaches that require training data orders of magnitude larger than the typical novel. In the middle ground, at the scale of the individual novel, a few machine learning techniques like Topic Modeling have proven useful for the literary scholar. However, Sentiment Analysis is unique in revealing latent structures and demonstrating emotional valence that correlates surprisingly well with human intuition. Because performance is statistical in nature and dependent upon the particular task and text, the gold standard—which we use in this paper—is agreement with human experts.

In the field of literature, investigations of underlying structure have been common. Vladimir Propp, Joseph Campbell and, more recently, Christopher Booker have popularized the notion of exploring commonalities across stories. For the most part their approaches, however varied, have shared a focus on investigating similarities of events, actions, and causality. More recently, scholars have turned to a different feature of narrative—affect. Patrick Colm Hogan describes what he terms an affective narratology from a cross-cultural perspective. Hogan hypothesizes that there are underlying cognitive-affective ways that narrative works. His method examines affect, rather than events and actions, to discern an underlying narrative arc. Literary scholars like Matthew Jockers began using lexical Sentiment Analysis of narrative with his open-source R package, Syuzhet.*R*. This new application represents the emotional arc of a narrative by measuring the changes of "emotional valence" across any narrative. It relies on sentiment lexicons that consist of words most commonly associated with a positive, neutral, or negative emotional valence along a numeric scale like $\{+1.0, 0, -1.0\}$. Sentiment analysis involves analyzing a string of text (e.g. an entire novel) and graphing a smoothed version of the sentiment calculated for successive units of text like sentences. The sentiment score for each sentence is the sum of positive and negative values of all words in the sentence found in the lexicon.

Even more recently, increasingly complex computational approaches have been developed by researchers like A.J. Reagan et al. In "The Emotional Arc of Stories are Dominated by Six Basic Shapes," the authors employ a larger lexicon and more through statistical modelling and smoothing methods to explore the freely available narratives on Gutenberg.org. They argue that the vast majority of narratives adhere to a limited number of emotional arcs. To verify their findings as meaningful, they take the same corpus of texts and generate two randomized versions of the text. These scrambled fictions, when passed through the same Sentiment Analysis, dissolve into a narrower band of noise compared to the distinct narrative patterns seen in the original text.

Jockers suggests more plots can be explained as subcategories of his basic forms—like branches on a family tree. More granular filters reveal these basic plots branching into more and more complex and discrete groups. Reagan et al. conclude that the vast percentage of texts adhere to basic plots while the remaining divergent plots comprise only a small percentage of texts. Scholars may employ different formulations, methodologies, and definitions, but all share a key focus on identifying emotional valence patterns in narrative that are shared across stories.

Recent state-of-the-art sentiment detection relies upon combinations of newer techniques like word embeddings, neural networks and built-in heuristics (Zhang et al.). In this paper we chose to test the simpler lexical approach because it is more widely available to a broad range of scholars. Lexical approaches like the one we employ here offer the advantage of simpler execution without sacrificing too much accuracy (Mohey El-Din Mohamed Hussein, D). While a few sentiments are likely misread, our approach, when applied to a large enough corpus, can be surprisingly effective, especially when compared to humans (Mozetic). The theoretical maximum for the accuracy of sentiment classification is defined by intra-rater agreement. That is, given both the ambiguity inherent in language and differences across individual perceptions, different humans assign the same sentiment values to training samples around 79% of the time. To complicate matters, however, individuals disagree with themselves over time.

Other studies use Cohen's Kappa or Krippendorf's Alpha to quantify intra-rater agreement among humans working on sentiment labeling tasks via Mechanical Turk or CrowdFlower, but all show similar concerns over the degree of intra-rater disagreement. These considerations must be kept in mind when comparing computational to human performance.

The Syuzhet.R package has the advantage that it has been verified by literary scholars through a comparison of readerly and computational sentiment values.[3] It also has a large lexicon and comes with a well-designed API for preparing text, customizing the analysis, and graphing. It has been subjected to public review by critics like Anne Swafford and most objections have been addressed (see appendix A). Still, we were concerned about the obvious shortcomings of it's naive lexical approach—a frequent criticism of this technique. Irony is one of several linguistic constructs that Sentiment Analysis systems, especially simple lexical techniques, struggle to accurately measure. A text for which irony provides one of the key aspects of interpretation would thus make a poor test case for computational analysis. Even if irony is not an overt aspect of the text, a closer inspection of inflection points should include a continued interrogation of the degree of irony in the passage in question.

Other challenges for Sentiment Analysis include slang, idioms, domain dependencies, world knowledge, subjectivity detection, and entity identification. The effects depend upon the statistical profiles of these features in the text and can vary across different domains (e.g. using a sentiment lexicon developed for movie reviews to analyze novels), which is why we benchmark simple lexical Sentiment Analysis with more complex approaches that include heuristics that deal with more common syntactic challenges. This provides some confidence that the models generally agree with each other as well as with the human literary expert (Baziotis). Specifically, we chose to compare the simple lexical Sentiment Analysis of Syuzhet.R with the popular library VADER that corrects for perhaps the most common shortcomings outlined above. While VADER does lexical Sentiment Analysis, it also comes with heuristics to detect and correct for common situations missed by pure lexical approaches including negations (e.g. "not happy"), intensifiers (e.g. "very happy"), contractions as negations (e.g. "wasn't very good"), and punctuation (e.g. "good!!!").

---

[3] For further information on Jockers and others creating labeled datasets see http://www.matthewjockers.net/2015/12/20/that-sentimental-feeling/.

The overlaid plots of sentence sentiment scores for *To The Lighthouse* using Syuzhet.R and VADER show very similar distributions with very similar means, variances and slight negative skews. Perhaps the biggest difference arises from the noticeably smaller areas of the VADER histogram except where it concentrates around neutral sentiment. This difference can be explained by the fact that VADER has half the lexicon size as Syuzhet.R (7,000 vs 14,000) so many more words would be labeled as "neutral"—the default value for words not contained in the lexicon. This result lends credence that the simpler lexical approach of Syuzhet.R is not statistically distorted by the common criticisms cited above, at least for our novel.

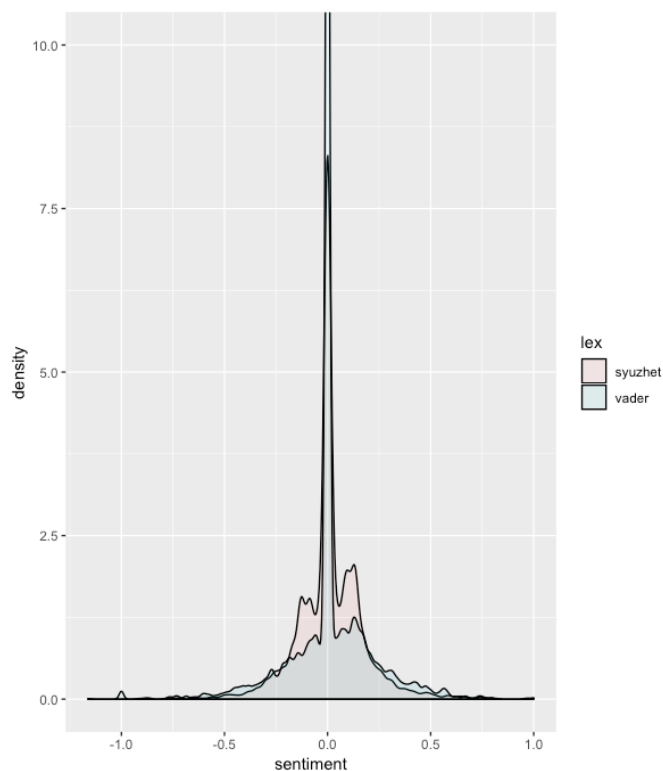

Figure A - Probability Densities of Sentence Sentiments for *To The Lighthouse*

First, however, one more note on methods and outcomes is in order. Computational approaches to literature are often criticized for offering no more than confirmation of the obvious. One could argue that Reagan et al. confirm what we already know—that the vast majority of stories are variations on the same few plots. The modernist novel, however, provides an interesting test case. Many would argue that significant shifts in narrative occur around the turn of the twentieth century. Woolf articulates this new development in "Modern Fiction": "If [a writer] could write what he chose, not what he must, if he could base his work upon his own feeling and not upon convention, there would be no plot, no comedy, no tragedy." "We are suggesting," Woolf writes, "that the proper stuff of fiction is a little other than custom would have us believe it." Here, Woolf contrasts the accepted style of convention with feeling and suggests that writing based on feeling would evince "no plot...in the accepted style."

Not long after she commented on this new development in fiction, Woolf wrote *To the Lighthouse*, a novel that appears relatively plotless when viewed through the action-event-causality lens mentioned earlier. The novel opens with the possibility of a trip to the lighthouse that takes place years later at the conclusion of the narrative. Echoing this delayed action, the artist Lily tries to finish a painting but completes the work only in the final section. Major events of the novel—like the death of key characters—are noted only briefly in the middle section, "Time Passes." Time—that great driver of plot—is less forefronted in two of the three sections, each of which takes place over a single day.

What if narratives that seem to have no plot reveal a strong emotional arc? Such a development would offer an interesting corrective to the way that we understand any individual modernist narrative and the developments of modernist fiction more generally. On the other hand, such surprising results would need to be correlated with the impression of so many critics that something significant has changed. It would also force us to ask more detailed questions about the relationship between plot and emotional arc.

Reagan et al. are careful to note that "the emotional arc of a story does not give us direct information about the plot or the intended meaning of the story, but rather exists as part of the whole narrative (e.g., an emotional arc showing a fall in sentiment throughout a story may arise from very different plot and structure combinations)." While Jockers emphasizes the term 'emotional valence' in lieu of 'emotional arc' or 'narrative arc,' he seems to concur with Reagan et al. that Sentiment Analysis informs but cannot be seen as a direct substitute for plot.

A closer investigation of Syuzhet.*R* and our test case text reveal certain methodological conclusions. First, selection of the text to be analyzed requires care and thought. Overtly ironic texts or texts with counterintuitive wordplay make poor objects of study for this particular tool. Second, a comparison with VADER should be employed to ascertain the prevalence of intensifiers, negations, and contractions that could be common enough to compromise the statistical accuracy of Syuzhet.R. Finally, one should consider whether the discovery of an emotional arc could lead to new insights about the text that would be of interest to the literary scholar. Texts that seem fairly straightforward in terms of emotional arc might be less appropriate for in-depth exploration.

**Comparing Models of *To the Lighthouse***

In spite of its seemingly plotless nature, Woolf's narrative exhibits a fairly strong emotional arc according to a variety of Sentiment Analysis models. *To the Lighthouse* appears to contain a distinctive latent emotional structure similar to those identified by Jockers and Reagan. Below is a graph with several different computational models and smoothing methods superimposed upon each other. How to choose which model to analyze? Here we stress a comparative and exploratory approach that analyzes similarities and differences between one example of smoothed emotional arcs from the two models: VADER vs Syuzhet.

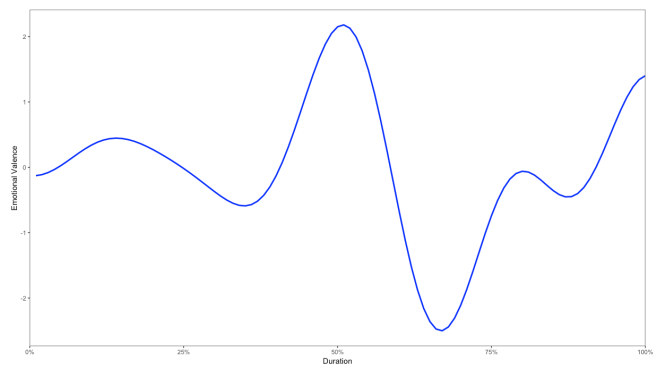

FIG B - VADER Sentiment Analysis of To The Lighthouse

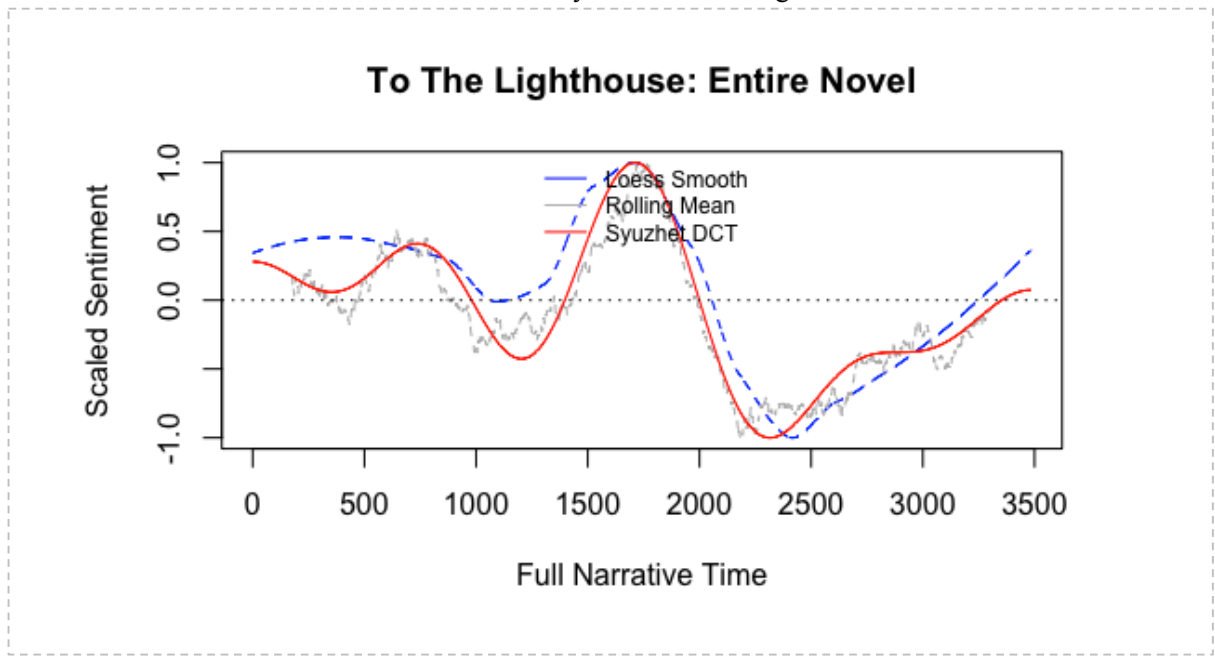

Figure C - Default Syuzhet.R Sentiment Analysis

Figure B shows the emotional arc using VADER. Figure C shows the emotional arc or valence using three different superimposed smoothing techniques: LOESS (LOcal regrESSion), Rolling Mean, and DCT (Discrete Cosine Transform). These represent four distinct approaches to extract a meaningful underlying signal from noisy data. LOESS is a nonparametric statistical approach that creates a curve by stitching together many small polynomial segments fitted to local data points. DCT is a parametric statistical approach that approximates a curve as the sum of a number of fundamental cosine waveforms. Finally, the Rolling Mean creates a curve by sliding a window of fixed size over the sentiment values for each sentence and assigning the average for the value at the center of the window (thus it is clipped at both the start and end).

What is immediately noticeable is how these very different models and smoothing methods are nearly uniform in agreement as to the general peaks and valleys of the emotional arc in *To the Lighthouse*. This holds true even though the methods are quite different. In this case, the method of the Rolling Mean and the parametric DCT produce fairly similar graphs. This graph also somewhat tallies with that produced by VADER.

This similar finding suggests that the overall shape may hold validity. At the same time, the extent to which three different smoothing methods, LOESS, the Rolling Mean, and VADER create divergent graphs for the opening of the narrative suggests that the beginning section in particular may pose challenges for plotting emotional arc. All of these possibilities will be explored further with the middle reading developed in the next section. For now, what is important to stress is that the careful consideration and comparison of various modelling and smoothing techniques gives us more information than a graph offering us only one choice.

Furthermore, one can adjust the smoothing filters of Syuzhet.*R* to view more generalized shapes or more granular detail. For the literary critic, we tend to assume more detail is always better, but in the case of Sentiment Analysis, a lack of filter prevents us from perceiving the general emotional arc. Below, for example, is a sentiment plot devoid of any smoothing filter.

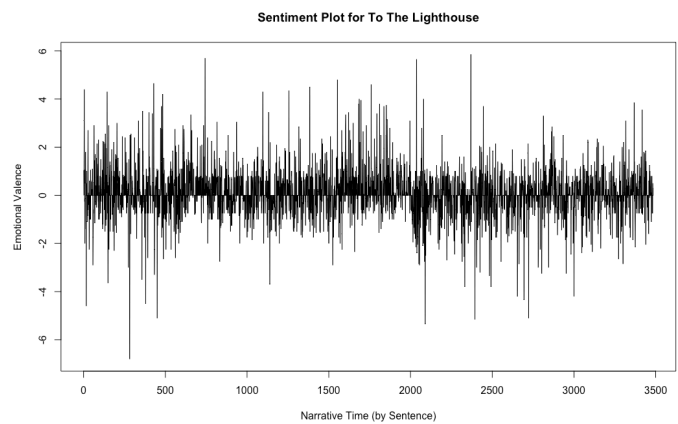

Figure D -- *to the Lighthouse* With No Filter

As we consider the various filters, of course, we can only hypothesize about the human reading experience. At this point in time, with our limited understanding of the cognitive experience of reading, we can only surmise that these underlying patterns, if shared across stories, must make up a significant aspect of the reading experience. Even if we accept this as true, it does not give us exact parameters for how to choose a filter that would offer the right compromise between detail of particular textual moments and apprehension of the more general arc beneath the detail.

The solid red line in Figure C represents DCT smoothing with a fairly high low pass value of 10. This results in a graph with more undulations. It may represent high-order information in more complex waveforms but distracts, perhaps, from seeing the fundamental rhythms driving sentiment. Jockers advocates the DCT method with a default low pass value of 5 in order to reveal the basic underlying shape. Figure E, labeled Simplified Macro Shape, shows the emotional valence of Woolf's *To The Lighthouse* using Jockers' recommendations.

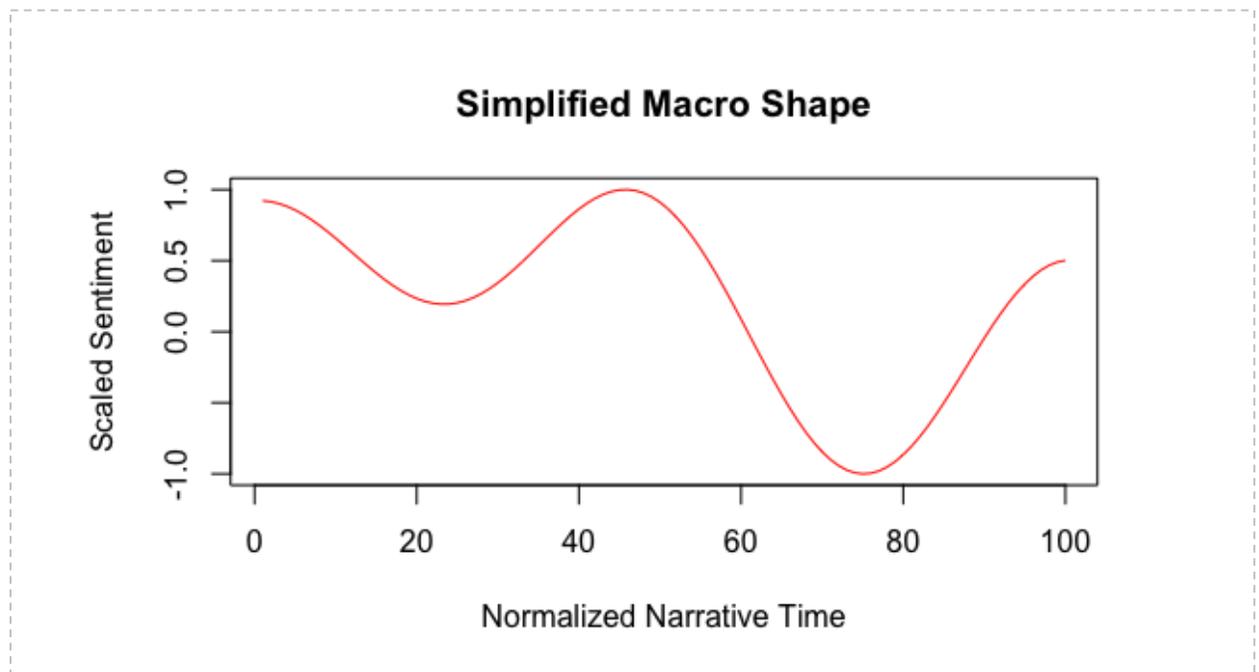

Figure E - Syuzhet.R sentiment plot of To The Lighthouse with DCT low pass filter = 5

The simplified shape echoes various aspects of the superimposed shapes in Figure C but offers slightly different information. Like the LOESS plot in the superimposed graph, the beginning of the narrative is smoothed into a single dip; like the DCT and Rolling Mean, the simplified shape starts with a dip rather than a hump. Paraphrasing the statistician George Box, all models are wrong, but some may prove useful. While this might be frustrating for a literary critic, who wonders why we should be exploring false models at all, it is worth pointing out that literary critics themselves explore fictional models of the world that they believe are crucial for obtaining insights about their world. Still, we need to assess the accuracy of these models their smoothed emotional arcs by some method. It's worth stressing that these computational Sentiment Analysis techniques rely on a statistical approach generally found to be more accurate on larger texts. They are not strong, in other words, in dealing with details or subtleties that rely on world knowledge or subjectivity detection. By contrast, this is often, one hopes, the strength of the literary critic. In the following section, we consider an approach that is neither close reading nor distant reading, but employs a middle reading[4] that pairs the graphs with readerly analysis.

**Analyzing the Rolling Mean Using Close Reading**

The Rolling Mean is commonly used to extract underlying long-term trends in stock prices and weather conditions masked by rapid short-term fluctuations. By plotting an average value over a sliding window of many data points (e.g. a week, month, quarter) daily swings can be smoothed revealing underlying long term trends. The Rolling Mean of *To the Lighthouse* results in a more jagged plot of sentiment than a DCT, which relies on detecting the underlying abstract pattern of a cosine wave. Nonetheless, the Rolling Mean largely comports with the arc given by DCT values but with an added level of detail that might aid further interpretative examination. Because the Rolling Mean of sentiment is contextualized, we can explore the surrounding context of each high and low point and gauge whether the computational mean approaches readerly experience.

---

[4] At present computational tools do not provide the same level of nuanced information regarding more traditional plot characteristics. The field is changing rapidly, however. See, for example, new approaches like Reiter et al. and Schmidt. Laurie N. Taylor also discusses the attempt to combine close and distant reading in a kind of middle reading for historical documents. See http://laurientaylor.org/2012/07/13/distant-reading-and-middle-reading/.

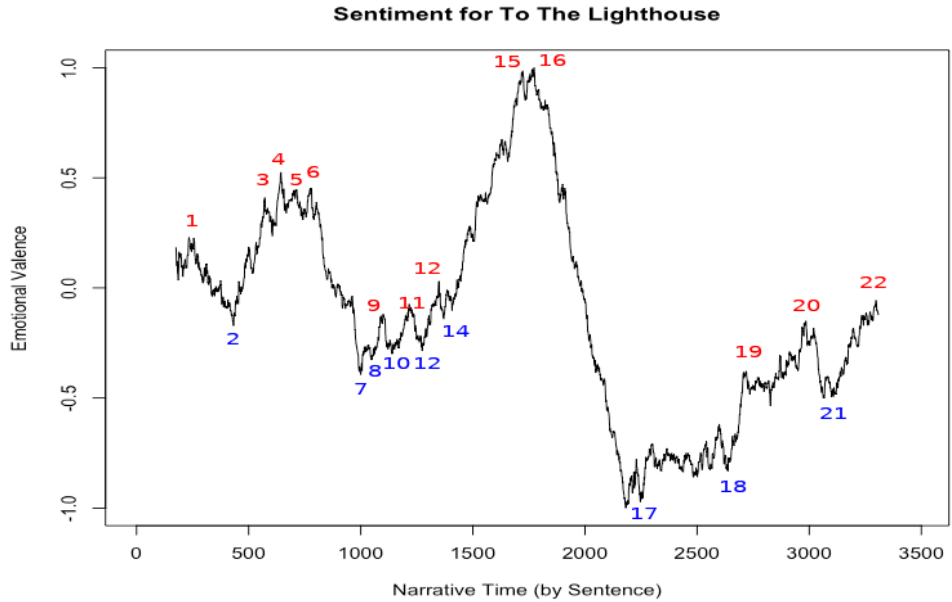

Figure F - Emotional Valence of "To The Lighthouse" (Rolling Mean, window 10%).

The first and last parts of the emotional valence plot are clipped by half the width of the sliding window which are 5% and 10% in our plots (i.e. the first sentiment value is calculated at the 5% point as the mean value for the midpoint of the 10% sliding window). Maxima and minima are entirely approximate based upon averages calculated over the entire window. This poses a methodological dilemma for literary criticism, which often relies on an exact "crux" to analyze. How might we analyze statistical points of inflection that are averaged rather than precise and generally more accurate over larger window sizes? One way would be to compare the surrounding text of the averaged "crux" to see if the contextualized inflection point seems to make sense. Even here, however, the human approach does not mimic the computational. The Rolling Mean window is quite a bit larger (i.e. 10% of the text) than what critics normally employ when judging context. Our analysis will compare a reader's contextual analysis on the smaller scale (approximately ten sentences on each side) with the larger contextual window. Since our contextual analysis adds another point of reference rather than simply replicating the computed emotional arc, we can bring one further literary technique to bear. One way to analyze the importance of a particular "crux" is to ascertain if it belongs to a larger thematic pattern of the text that helps illuminate meaning. Here, therefore, we bring two aspects of close reading to bear by asking a) does the statistically-determined sentiment accurately reflect the sentiment of the inflection point and b) does it locate inflection points that when viewed in aggregate provide a thematic pattern of the text?

For brevity's sake we do not provide the entire context for each inflection point. Readers are encouraged, however, to consult the text and replicate our findings. Readers may consult Figure F to ascertain where points fall on the emotional arc. The narrative first rises as Mrs. Ramsay and James hold hands and leave the dining room together (P1). The arc then dips when Mr. Bankes explains to Lily that "the number of men who make a definite contribution to anything whatsoever is very small" (P2). The emotional arc rises again to a positive peak in a series of highs that include Mrs. Ramsay smiling as she regards the transition of Mr. Ramsay's thoughts from negative to positive, and Mr. Ramsay "safe, restored to his privacy" as he begins to engage his "splendid mind"(P3). The highest trajectory of the arc occurs as Mr. Ramsay imagines the need "first to be assured of his genius," and "then to be taken within the circle of life, warmed and soothed, to have his sense returned to him," to have a house "filled with life"(P4). The next peak occurs when Mr. Ramsay pauses to look at Mrs. Ramsay reading to James. He looks, nods, and approves of what he sees (P5). The end of this high multi-peaked curve ends with the moment the characters "became part of that unreal but penetrating and exciting universe which is the world seen through the eyes of love"(P6).

Let's analyze this emotional arc more closely. First, the averaged points do comport with readerly assessment in terms of emotional value. But do they reveal a thematic pattern? The emotional low point concerning the small "number of men who make a definite contribution to anything" provides an interesting thematic thread that is counterbalanced by the contributions of the many women in the novel. Here we might include the work of Mrs. Ramsay, the cook, and the daughter (crafting the artistic centerpiece), each of whom contributes to the emotional high point of the dinner discussed in more detail in a moment. We might also add the work of the cleaners during "Time Passing" and Lily's painting, which, Woolf writes, may never be seen or noticed.

We could say that the emotional low point typified by skepticism concerning mankind's contributions is thus opposed, at an emotional peak, by a complementary understanding of two very different kinds of contribution. The high point of the narrative couples Mr. Ramsay's notion of "genius" with the importance of Mrs. Ramsay's contribution in creating a house "filled with life" that provides warmth and sense. This complementarity resonates with Virginia Woolf's conclusions in *A Room of One's Own* concerning the power of conjoining so-called "male" and "female" traits. At this point, therefore, the emotional arc seems consistent not just in terms of emotional valence but in terms of making manifest underlying thematic patterns.

Perhaps a trickier question concerns the more subtle understanding of the character's thoughts. How do we readers experience Mr. Ramsay's emotions given the clues we get as to the flaws in his character? While not irony, many literary critics might well argue that we cannot take many of the sentiments at face value even when represented as free indirect discourse in the thoughts of a character. Instead they are filtered through our understanding of character. If we hold to this view, the emotional arc provided provides a very different view of the narrative than that provided by readerly analysis.

To rebut this view, however, we might point out that many who teach literature believe it allows us to step into the perspective of a way of thinking or being quite different from our own. When discussing a character we can easily ascertain limitations, therefore, while still taking seriously the emotional experience of that character. This second view holds more in common with recent neuroscientific findings about the mirror quality of emotional empathic experience. While we don't discount the necessity of ascertaining complex views of fictional characters, therefore, we still argue for the importance of investigating the raw emotional arc of the narrative. If the novel makes available "transparent minds," as the critic Dorrit Cohn formulates it, then Syuzhet.R can offer us a view into how those transparent minds experience emotion.

The next section of the Rolling Mean offers yet more methodological considerations, as the graph represents a jagged up and down movement that never rises above neutral. Do we read the dips and rises (coded in blue and red) as opposite because of their relative difference, or as different shades of negative emotion. Let's take a closer look to see whether a close reading can help shape our interpretation. First, the arc takes a downward turn, moving fairly rapidly to below the neutral line with another series of jagged peaks that include the moment when Mr. Ramsay "was angry" (P7) after Mrs. Ramsay asks why the children have to grow up, followed by the moment when Mrs. Ramsay retreats into herself, away from all the "being and doing" to a "wedge-shaped core of darkness"(P8), The next low point finds Mr. Ramsay wishing "Andrew could be induced to work harder"(P10)—wishing him, in other words, to engage in the task of trying to be one of the very small number of men "who contribute anything whatsoever." What may surprise the reader is that these low moments seem to surround underlying tensions between characters, as for example, between Mr. Ramsay and his focus on "being and doing" in contrast to Mrs. Ramsay's entirely different set of values and concerns. They do therefore comport with readerly interpretation both on the valence and on the thematic level.

    In contrast to these low points are rising high points that reach toward neutral. The first (P9) finds Mr. Ramsay letting Mrs. Ramsay be. The point remains slightly negative however, for at the same time as it reveals a touching moment of intimate understanding, it "hurt him that he should look so distant and he could not reach her, he could no nothing to help her." Next and rising to near neutral, Mrs. Ramsay thinks "with delight how strong [Mr. Ramsay] still was..and how untamed and optimistic" (P11). While this emotion of delight is clearly positive, it does serve as a contrast to some of the other thoughts that contextualize the moment—Mrs. Ramsay's annoyance at the "phrase-making" and the "melancholy things" Mr. Ramsay says, as well as "the horrors" that seem not to depress but to cheer him. The arc then dips once again to the moment when Prue reaches up to catch a ball while "darting backwards over the vast space (for it seemed as if solidity had vanished)"(P12). Complementarity and appreciation are linked with the emotional rise. A vision of things and people separated by great distances, "blown apart," leads to a low and echoes other negatively charged moments in the text—both the earlier tensions that separate characters and the effect of time passing that, with the war, creates separation and distance.

    Similar to the moment when Mr. Ramsay lets Mrs. Ramsay be, we find the next point, P13, hovers around neutral because there is a pronounced ambivalence of emotions that counter each other.

Even as Paul and Minta return from their walk having agreed to marry, Minta is "crying for something else" (P13). Redescending below neutral, again Mrs. Ramsay feels sadness at the thought that "it was so inadequate, what one could give in return to her daughter, Rose," and that "she would suffer" (P14). These complex emotional valences suggest that human connections and relationships are not enough to counter the sorrows of life even as they provide a contrastive balance that brings the points toward neutral.

All of these emotional valences seem to correspond to an underlying thematic pattern. Negative emotional valence seems to occur around the moments in which relations between characters are strained or questioned: between Mr. and Mrs. Ramsay, between the spectators watching Cam's ball, and between Mr. Ramsay and Mrs. Ramsay and their children. There is a relative movement toward neutral, by contrast, when there are conflicting emotions that seem to balance each other and provide a relative rise in emotional state—in the example of Minta's mixed emotions at her engagement, for example. Finally, very positive emotions are graphed when there are complementary balances or connections between selves—with Mr. Ramsay's appreciation of the house full of life that marriage has made possible, for example.

Once we expand beyond a very small readerly contextual window, however, analysis becomes even more complicated. The moment of P5 finds Mr. Ramsay feeling positive emotions when gazing at his wife and child, but Mrs. Ramsay's emotions are a bit more complicated: "Mrs. Ramsay could have wished that her husband had not chosen that moment to stop." Similarly, the negative moment surrounding Cam's catch of the ball is framed on either side by more positive emotions. Immediately before, Mr. Bankes has agreed to dine ("it was a triumph") and immediately following it Mr. Ramsay "felt free to laugh" thinking of Hume stuck in a bog.

This vacillation between positive and negative emotional valence is prevalent in the opening section of the narrative. When Mr. Ramsay leaves Mrs. Ramsay alone, it is because he sees her as very distant. His action is motivated by love but performed with a slight sense of dissatisfaction. However, this moment is preceded by Mrs. Ramsay's experience of the lighthouse (likely the cause of her distance), during which she experiences intense joy as she thinks "It is enough!" Is it really so negative when Mrs. Ramsay experiences herself as a dark-shaped wedge, since she can go anywhere and be anyone? When she has a moment's respite from "all the being and doing" required of her as a mother and wife, and her husband decides to "let her be," we can theorize that these are emotionally ambivalent moments.

Other critics have focused on the vagueness and ambiguity in this succession of emotions.[5] But one might instead suggest that we have a series of impressions that lead to powerful ambivalent emotions. These ambivalent moments mirror, in point of fact, Lily's more poetic impressions in this first section of the novel. For Lily each character is thought of as a composite of positive and negative traits, and her relation to them is thus emotionally complex. For example, Lily tries to make sense of her conflicting impressions of Mr Bankes, who is both a "generous, pure-hearted, heroic man," at the same time as "he objects to dogs on chairs and the iniquity of English cooks." It is also worth noting that the readerly response to the novel may mirror this experience. As an example, Mrs. Ramsay's positive emotions about Mr. Ramsay may be countered by the conflicting less positive thoughts of other characters.

One might be tempted to conclude that *To the Lighthouse* provides a test case that is far too complex to capture with Sentiment Analysis techniques. However one could also conclude the contrary because the statistical means do still seem to comport with readerly analysis. To some extent, this jagged alternation of highs and lows is mirrored on a larger scale by the Rolling Mean graph. However, it is probably useful to bring in the other techniques for further comparison and examination. In general, when working with noisy data, one can feel more confident in one's conclusions when two very different approaches provide the same information. In our case, it is this first part of the narrative that demonstrates the most variability according to which approach is chosen.

---

[5] See Molly Hite's *Woolf's Ambiguities* and Megan Quigley's *Modernist Fiction and Vagueness*. While both are right to stress ways in which the narrative resists reading for meaning as a fixed, stable interpretation, they overlook the extent to which the emotions are not vague or ambiguous but rather bi-valent, much as the larger narrative is.

The opening section offers the most varied arcs between models, and the various smoothings—levels of filter—further demonstrate different ways of extracting the underlying emotional arc. DCT smoothing reveals the opening narrative to be defined by a small dip that precedes the two larger "man-in-a-hole" dips later. This largely comports with the Rolling Mean, although the Rolling Mean shows that there is far more variability in values. The LOESS graph, however, smoothes the opening of the narrative into a "hump" rather than a "dip" in emotional valence.

Given the variability between the graphs in this opening section, we might conclude that the beginning of the narrative might be the most difficult to plot. While many critics focus on issues of negation, qualification, and irony, therefore, the larger issue may be whether some moments are emotionally contradictory and/or ambivalent. Furthermore, an analysis of the Rolling Mean does suggest that one gets a sense of relative emotional states in which "neutral" can actually be interpreted as ambivalent. Many other composite, ambivalent thoughts also occur in this beginning section of the narrative and would give credence to a reading of the opening section as one in which emotional arc might be complicated by Woolf's exploration of a less conventional approach to narrative through the use of poetic impressions—a point bolstered by Woolf's own discussions of modern fiction. In some ways, Woolf's novel theorizes just such impressions and provides an excellent case study for how to use (or misuse) Sentiment Analysis.

Are we meant to focus on poetic impressions, both our own and that of the characters, which are contradictory and often highly ambivalent? Or on the underlying arc? Perhaps both give us the most complex and nuanced understanding of the narrative. While the Rolling Mean does seem to provide some sense of an underlying emotional arc based on certain thematic parallels, it also seems to be the case that the readerly analysis and fairly dramatic differences between smoothing techniques reveal an emotional arc that is somewhat noisy in the beginning—a phenomenon that is overtly represented in the narrative via the thoughts of Lily. Our comparison of readerly contextual analysis with multiple techniques helps bring out the emotional complexity of the opening and shows that both our readerly interpretation and the variations between different approaches can reinforce each other.

The situation is quite different with the next highs and lows of the Rolling Mean. The various smoothing techniques converge mid-way through the narrative. At this point in the narrative, the DCT with a filter of 10 (overlaid in Figure A) and the Rolling Mean appear remarkably similar. We can surmise, therefore, that the emotional arc becomes stronger and easier to discern. We continue with the Rolling Mean analysis with the understanding that this approach has the added value of comporting in a general fashion with the others. The focus here will therefore be on whether and how it might compare with reader experience.

The high point of the narrative occurs almost half-way into the narrative, during the dinner party. The first of two peaks occurs when Mrs. Ramsay orders the children to light the candles and "the flames stood upright and drew with them into visibility the long table entire, and in the middle a yellow and purple dish of fruit" (P15). The second emotional peak occurs almost immediately afterwards, when Mrs. Ramsay thinks "We"—that was enough...They'll say that all their lives, she thought"(P16). The lighting of the candles provides a fairly stark contrast with the earlier emotional low linked to a loss of solidity, distance, and separation. The table centerpiece and marvelous stew that are described immediately following the lighting of the candles also both invoke images of coherence during which very different elements are brought into relation. Finally, we have a sense of character connection, not just between the various diners but between Minta and Paul. It is an affirmation of 'we' that will endure—"'they'll say that all their lives,' thinks Mrs. Ramsay."

Next, our graphs are fairly similar in suggesting a dip in emotional valence. The Rolling Mean contextualized point occurs when "one by one the lamps were all extinguished"(P17)—a direct contrast to the lighting of the candles. It is a descent into darkness and chaos that counters the coherence that emerges during the dinner party. Mirroring the moment when Cam catches the ball, we have a loss of solidity as night descends and, as Prue observes, "One can hardly tell which is the sea and which is the land." Relations between characters become distanced, and Woolf writes that "somebody laughed aloud as if sharing a joke with nothingness."

This nadir is followed by several additional low points that occur in the third section. First, Lily reflects on her distance from Mr. Ramsay: "He had become a very distinguished, elderly man, who had no need of her whatsoever" (P18). Then Lily thinks about Charles Tansley, who insisted that women can't paint and always impeded her views—first, she remembers, while she was painting and then, while she was eating (P19). While the emotion is negative, there is still a rise from points 17 and 18, perhaps because the negative comments of Charles Tansley are offset by the more positive experience of an artist exploring the beauty of the world around her. The emotional arc of the narrative rises still further while remaining below neutral, with the imagined vision of Paul and Minta finding common agreement in the moment when "he sat on the road mending the car, and it was the way she gave him the tool—business-like, straightforward, friendly, that proved it was all right now"(P20).

A dip occurs again with Lily's frustration at her inability to communicate to Mr. Carmichael or to express herself with words. Here the low point is tied to an experience of emptiness and nothingness that echoes the low point of "Time Passing" and brings to Lily's mind the conflict between "wanting and not having"(P21). The final rise of the story ends with Lily's thought that at certain moments one "felt something emerge." The emptiness changes to fullness: "Empty it was not, but full to the brim." At these points, Woolf writes, "Life [is] most vivid" because "some common feeling...held the whole together"(P22). Lily experiences this as a "feeling of completeness" that she had first experienced when gazing on Mr. and Mrs. Ramsay during the earlier trip to the summer house.

The points of inflection seem to correlate with an analysis on both the micro and macro level. That is, the emotional valence, when vetted by a closer look at the surrounding context, seems to comport. The various moments also seem to reflect key points of the text that other scholars find significant. For example, Stephen Kern argues that *To the Lighthouse* reflects a thematic concern with fragmentation and unification that echoes our thematic pattern here of separation and connection, dissolution and emergence. The passages he chooses to demonstrate this pattern also dovetail with those highlighted by Sentiment Analysis. To further support Kern's argument, he quotes Virginia Woolf, who many times refers to her desire to find the pattern beneath the chaos of life. Explaining her process when composing *Voyage Out*,

Woolf writes of her attempt "to give the feeling of a vast tumult of life, as various and disorderly as possible . . . and the whole was to have a sort of pattern, and be somehow controlled"(cited by Kern, 317)

Sentiment analysis reveals this concern with chaos and pattern to be a continuing preoccupation of Woolf's. Indeed, the emotional arc of the novel mirrors this same gradual coherence of shape out of chaos as the arc moves from less coherent emotional valence to more. While giving us a slightly more detailed picture of how this thematic pattern weaves through the entire narrative, Sentiment Analysis also provides a way to understand why certain passages are not assessed as points of inflection. Kern, for example, cites two passages that are not highlighted by our model, and exploring why provides further insight into what Sentiment Analysis offers as a descriptive tool.

The first passage occurs after the dinner party when Mrs. Ramsay refuses to express her love in a way that Mr. Ramsay would like to hear. The couple then proceed to read separately, finding solace in entirely different literary experiences—Mrs. Ramsay, through reading lyric that has a more complex emotional valence and Mr. Ramsay, in his reading of a Romantic novel. Why is this passage "lower" on the emotional valence scale than the dinner party? Perhaps because it provides a complex and ambivalent portrayal of a moment of intimacy.

If we return to the underlying thematic pattern of our earlier points of inflection, we can see that when characters and the surrounding world seem to scatter or experience the world individually, the emotional arc moves downward. When people and objects cohere in an emergent state of connectedness, the emotion rises to a high. Emotional lows and highs also seem to reflect when two people don't will the same (Mr. and Mrs. Ramsay, Charles Tansley and Lily) or when they do (Mr. Bankes commenting on the triumph of Mrs. Ramsay's dinner, Paul and Minta saying "we"). Sentiment analysis thus gives us a way to understand the moment of intimacy shared by Mr. and Mrs. Ramsay as one in which a certain amount of conflict and tension is reflected in a lower emotional valence than the high point of the dinner party.

Another significant passage occurs in "Time Passes" when the death of Mrs. Ramsay is described in brackets. It would be easy to argue that Sentiment Analysis fails entirely when it comes to graphing this moment—so emotionally wrenching for the reader—since the language is devoid of emotional lexical indicators. However, one can also see that this moment, while devoid of any emotional vocabulary, is surrounded on both sides by the emotional "hole" into which the emotional arc descends. The sentence might be compared to a moment of silence or pause that only further accentuates—rather than contradicts—the intense negative emotion of the surrounding text. Again, Sentiment Analysis allows us to predict the emotional valence of this moment; this time, we can predict that this moment should fall (as it does) during the lowest emotional period of the novel. So far, our analysis both comports with and provides deeper insight into aspects of the novel first discovered using traditional literary methods of interpretation.

We return, then, to our initial question: how useful is this Rolling Mean? These cruxes do seem to comport generally with emotional highs and lows, although the first moments are embedded within a narrative that contains many contradictory emotional moments. At least one further question remains to be asked and, at least provisionally, answered. A 10% Rolling Mean window is very large—approximately 350 sentences. Is it possible that we might be able to justify *any* underlying pattern that the computational analysis "discovers"? Perhaps one could create a meaningful interpretation using just about any points? While more exploration surrounding this question is needed, we conducted two further experiments.

In the first, inspired by the methodology of Reagan, et. al., we generated ten randomized word "salads" of the novel and compared the resulting sentiment plots to the original novel. The sentiment plot of the original novel clearly stands out from the noisy band containing the ten randomized plots. This finding suggests an organized latent and/or intentional structure absent from the word salad versions of the novel.

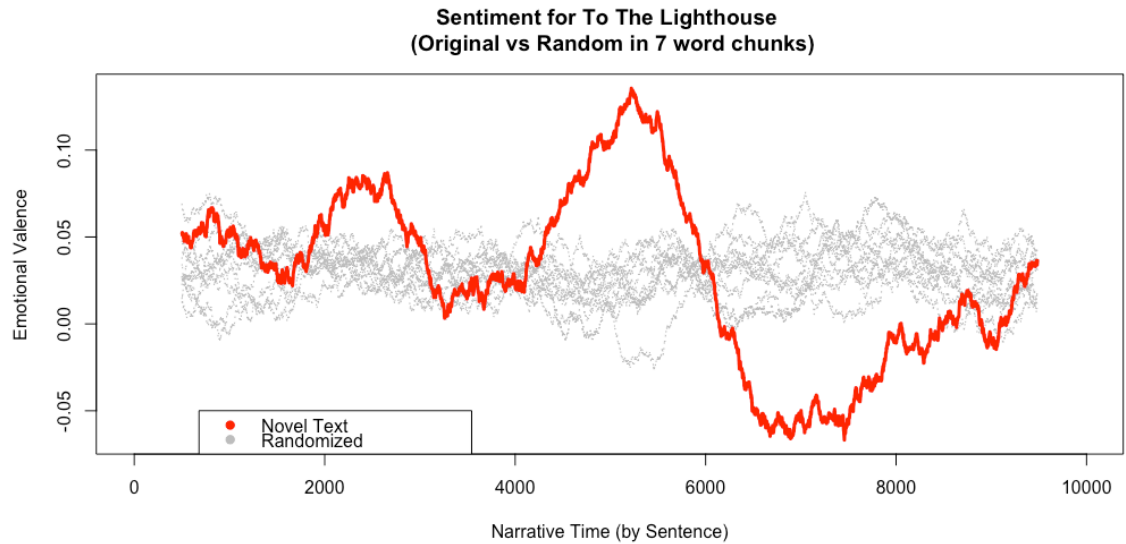

Figure G

We also explored the possibility of using a smaller rolling window which, we hypothesized, might approximate more closely an attentive reader's examination of context. At 10% the default windows size is 350 sentences, but smaller windows are increasingly likely to mislead our statistical approaches to Sentiment Analysis. What size window to choose? When the window size becomes too small the plot becomes more jagged and noisy, thereby rendering the underlying arc more difficult to discern. The optimal Rolling Mean window thus seems to be one that balances granularity of information with smoothed noise to make visible the underlying arc. At 1% and 2.5% the plots seems to provide too much background noise, but would 5% be a better window to employ by offering the best compromise? To test this hypothesis, we did another analysis of the inflection points using a 5% Rolling Mean.

Briefly, the cruxes that surfaced with a 5% Rolling Mean were far more difficult to harmonize with human interpretation. As just one example, one of the higher emotional valences occurs when Mr. Bankes exhibits negative existential musings at the dinner party—the exact opposite of what we would expect. At this moment he thinks that everyone drifts apart and wonders what it was for, asking "Are we attractive as a species?" There are a few points that make more sense and in general, they comport with the shared moments of our 10% window—as just one example, the moment when Paul and Minta fix the car.

Interestingly, the global maximum and minimum occur in slightly different places than in our other plots. The global maximum occurs when Mrs. Ramsay thinks that Lily and Mr. Bankes should marry. The global minimum occurs when Mrs. Ramsay thinks later that Lily and Mr. Bankes are not in love but should still marry. This finer granularity turns the narrative into a failed love/marriage plot whose main characters are Mr. Bankes and Lily. While interesting (it may indeed be a sub-plot of the novel), the reduction of the narrative to these key inflection points fails to take into account the larger cast of characters or the major thematic concerns of the plot.

With the 10% window, by contrast, the emotional arc of the novel is distributed across various characters—a technique we call the "distributed hero(ine)."[6] Both emotional high and low points are seen through different perspectives. While it might be tempting to see the novel as poetic impressions of various minds without a unifying coherence, therefore, Sentiment Analysis allows us to see that there is an underlying structure that emerges mid-way through the novel. The many characters share part of a larger emotional arc. This phenomenon is mirrored in the plot in more metaphorical ways, because many of the highest and lowest points occur when perspectives converge and seem to amplify sentiment. While this might seem evidence of what Philip Petit has termed the "common mind," the distributed hero(ine)" actually suggests something a bit different. Moments in which the various perspectives seem to amplify emotions nonetheless display a distributed, reinforcing set of emotions that is all the more powerful for being distributed yet connected. Between these inflection points, moreover, we can see the emotional values that together comprise the arc of the story moving between characters in a way that maintains individual perspective while suggesting communication and dependence.

---

[6] Here we borrow from the computational notion of a distributed network. A distributed network occurs when the programming, the software, and the data to be worked on are spread out across more than one computer, but they communicate, or are dependent upon each other. Caserio comes close to describing this phenomenon when he notes of modernist narrative that a strict familial hierarchy and linearity is replaced by mutuality and adjacency of parts. Susan Stanford Friedman, in "Lyric Subversions of Narrative," opts for reading Woolf's novel not as the simple binary opposition of feminine lyric and masculine plot, but as a complex interchange of stasis and change that blends the relational emphasis of Mrs. Ramsay and the linear outlook of Mr. Ramsay.

In this way, the sentiment of the so-called hero—note the usual terminology of "man on the mountain" and "man in the hole"—is often distributed in Woolf's novel over many characters. It would be easy to see the modernist narrative as one in which multiple perspectives lead to fragmentation, however Sentiment Analysis allows us to perceive something quite different. While the narrative continually shifts between viewpoints, the emotions of these different characters together create a unified emotional arc. This phenomenon has implications for our understanding of the modernist novel, a point to which we will return in a future paper. For here, it's important to stress that while the Rolling Mean with a 10% window captures this nicely, the 5% window does not.

To conclude our analysis of various Rolling Mean windows, we suggest that, for the most part, the majority of inflection points found using the 5% Rolling Mean fail to offer a coherent reading of the novel and are often contradictory, as in the example with Mr. Bankes. While this may surprise the traditional literary scholar, for whom more detail is almost always better, it is less surprising when one remembers that statistical computational models do less well on smaller subsets of text. Further investigation is needed, but it does seem that a 10% window may offer us an experience of the narrative arc that aligns more closely with readerly experience. While this is true of Woolf's novel here, however, we should stress that similar comparisons should be undertaken to assess the accuracy of the various parameters when using Syuzhet.R for any textual analysis. The larger methodological conclusion is that comparison of various models and smoothings is necessary, at least until we have a larger dataset of sample cases to compare.

**A Closer Look at the DCT**

The DCT with a low pass filter (LPS) of 10 looks remarkably like the Rolling Mean with a window of 10% that we just analyzed. But what of the simplified shape using a filter of 5 that Jockers advocates (represented in Figure C)? While this kind of simplified shape is valued by Jockers because it allows for comparisons between narratives, does it offer the literary critic any new perspectives or insights?

The simplified arc graphs the beginning of the novel, which the Rolling Mean is unable to ascertain for reasons already discussed. The narrative begins with an emotional high point during which James imagines that he will finally be able to go to the lighthouse. The simplified macro shape then dips to a local minimum at the moment when Mrs. Ramsay reads and thinks "quite easily, both at the same time," the story of *The Fisherman and the Sea* alongside thoughts of Paul and Minta and their possible engagement. The fisherman tells the flounder that his wife "Wills not as I'd have her will." This conflict resonates thematically with the uncertainty of wills between Paul and Minta ("Will they marry?"). It also tells the story of a marriage in which husband and wife fail to balance each other out, not only willing differently, but embarking on a path of greed and "wanting but not having" that echoes our earlier thematic pattern.

This high point and dip nonetheless provide new information, since they foreground a narrative in which the anticipation that a particular quest—a trip to the lighthouse—might fulfill so much desire is counterbalanced by the finding, in the tale of T*he Fisherman and the Sea*, that reaching material goals often fails to bring about lasting happiness. While this narrative arc seems a bit different from the more granular one analyzed using the Rolling Mean, it does seem to confirm that singular and individual ambition is often coupled with disappointment, and it coheres with Mrs. Ramsay's thoughts, counter to Mr. Ramsay's quest to arrive "at the letter Z" in philosophy, that particular moments of connection are "enough."

The DCT LPF 10 and 5 show remarkable similarity with the global maximum and minimum sentiment of the Rolling Mean, thereby amplifying the sense that this underlying pattern is worth taking seriously. The simplified shape locates the maximum at almost the exact halfway point in the novel—at 46% of the narrative. Confirming our analysis with the Rolling Mean, it occurs during the moment when Mrs. Ramsay thinks, "that was enough" upon hearing the difficulty with which Paul for the "first time had said "we." In contrast to the low point of *The Fisherman and the Sea* tale, this moment establishes a relationship or momentary harmony between two people united in a connected experience. The DCT LPF

5 high point also forefronts the most conventional point in the novel in which two characters decide to marry.

The simplified DCT shape marks a slightly different dip in the narrative from the Rolling Mean, however or DCT with filter of 10 compared earlier. The emotional nadir occurs when Cam looks back at the shore as they sail towards the lighthouse and she cannot make out the summer house. The house has disappeared for the first time in the narrative, and the "shore seemed refined, far way, unreal." Distance has changed Cam's relation to the house: "the little distance they had sailed had...given it the changed look, the composed look, of something receding in which one has no longer any part." Again, separation is stressed and distance is key. (12)

The DCT also gives us a slightly different second peak, with a local max of .34 that is not nearly as high as the dinner party but still fairly positive. It occurs at the moment when "with a sudden intensity, as if [Lily] saw it clear for a second, she drew a line there, in the center. It was done; it was finished. Yes, she thought, laying down her brush in extreme fatigue, I have had my vision." For the most part, the valence of these highs and lows as well as their thematic patterns seem to comport with readerly analysis. While there is a steep rise at the end of the novel in absolute terms, the rise began from a very low point. The final emotional valence thus ends lower than the earlier dinner party. In terms of the novel, we can say that while the characters reach some more positive emotional state after the death of Mrs. Ramsay, it is never as positive as before her death.

What is interesting about the DCT simplified shape is that adds a slightly more distant reading of the novel while at the same time revealing an underlying pattern that mirrors that discussed earlier. A more distant middle reading reveals a thematic pattern that also, rather surprisingly, shows distance of vision as a significant way of understanding the world (both Cam's and Lily's). At the same time, the emotional high points occur when this distance is negated—either through marriage and the dinner party or through the visit to the lighthouse and the completion of a painting. Finally, this more general pattern highlights the significance of the most bare-boned of stories—the tale Mrs. Ramsay reads her son James—as an important method to understand both the novel and the world.

**Conclusion**

Our hybrid model offers us insights into both Syuzhet.*R* and *To the Lighthouse* that we will briefly summarize here. We start with our conclusions about Sentiment Analysis. In addition to carefully considering the text to be analyzed, critics should also investigate the various methods of modelling the emotional arc of the narrative. Special attention needs to be taken with regard to parameterizing the smoothing technique to extract the most meaningful sentiment arc (e.g. via window size for Rolling Mean and low pass filter value for DCT). This approach includes comparing VADER to Syuzhet.*R* and comparing and contrasting the various methods for ascertaining the latent emotional arc. We can hypothesize that where the emotional arcs agree, we find the highest probability that the "common reader" (to repurpose Woolf's excellent formulation) will agree. Where the models and smoothed emotional arcs diverge, more investigation needs to be done to determine what this divergence signifies in terms of emotional valence. Both absolute emotional valence and relative emotional valence also need to be taken into consideration, with the possibility that ambivalence or contradictory emotions can often result in a "neutral" plotting. Perhaps most importantly, the various approaches should be compared not only with each other but with readerly analysis to ascertain whether they comport. Here, we found that not only did they largely comport, but that they revealed thematic patterns that are significant to the novel.

On the one hand, Sentiment Analysis confirms what scholars like Stephen Kern have already ascertained about *To the Lighthouse*. At the same time, this method gives us a slightly different vocabulary to explain this pattern—one that relies on networked connections and adds a new dimension to Kern's argument about fragmentation and unity using traditional methods by suggesting connectivity and separation might be key terms. As we move through the various characters in a seemingly fragmented manner, the individual emotions reveal an underlying pattern. Moreover, as the story progresses, there is a movement from less coherence to greater coherence as a more pronounced emotional arc emerges. This emergence of a more coherent narrative in fact mirrors the relational coherence established by the characters during the dinner party. While the beginning of the novel may seem more ambivalent and emotionally complex, therefore, a strong emotional arc emerges as the novel unfolds.

Taking a more distant view of Woolf's novel, we might also interrogate its place in the new development of modern fiction at that time. While the modernist novel may hold to no conventional type, the various laments about the disintegration of plot (or conversely, scholarly exuberance over this development) may warrant further exploration when we understand plot to include, while not being limited to, sentiment arc. Like Kafka's invocation of more traditional stories, for example, or James Joyce's retelling of the *Odyssey*, Woolf's invocation of the *Fisherman and the Sea* reveals the extent to which these more traditional stories may provide underlying patterns and counterpoints and not just superficial invocations. Furthermore, before we dismiss the "distant" reading that this computational method provides, we should remember the reflections of Woolf's artist, Lily: "So much depends then, thought Lily Briscoe, upon distance: whether people are near us or far from us." The emotional arc of *To the Lighthouse* reveals the highs and lows in which "distance is key." Our readerly experience can counter this distance by helping us to take a closer look.

# Appendix A

## Addressing Critiques of Syuzhet.R

This paper uses the Syuzhet.R software package by Matthew Jockers[7] to perform standard lexical Sentiment Analysis. We chose this package for a number of reasons enumerated in the body of our paper, but also because it has been subject to public review for several years and validated against human experts. Part of this validation includes our own experience verifying that results from Syuzhet.R comport to the judgment of human experts.

In June of 2014[8] Jockers outlined his approach to analyzing plot using the evolution of emotional valence over the course of a novel (inspired by Kurt Vonnegut's failed master's thesis). Jockers released the first version of his code on GitHub in February 2015.[9] Within a month, Annie Swafford offered three specific criticisms of Syuzhet.R.[10] Her criticisms nicely summarize the main concerns with Jockers' Syuzhet.R and are addressed here in the order of seriousness.

**Objection #1: Discrete Fourier Transform Smoothing**

The main and most persistent criticism raised by Swafford was Jockers' initial use of Discrete Fourier Transforms (DFT) as one of several options for curve smoothing the emotional valence time series. This criticism was echoed by other scholars like Ben Schmidt[11] and resulted in a series of public and private communications between Jockers, Swafford and others to resolve this issue.

The fundamental problem using DFT for curve smoothing was that it is designed to model periodic signals and thus imposes a boundary condition that both starting and ending points have the same value (so the signal can be repeated in a continuous fashion). This had the unfortunate and obvious effect of distorting the emotional valence curves at the start and end of the novel.

In April 2016,[12] after over a year of critique, testing and consultations with experts in signal processing, Jockers discovered two solutions to the edge distortions introduced by DFT smoothing and released an updated version of Syuzhet.R. The first solution allowed users to add a variable length pad before and after the emotional valence time series to minimize the effects of curve distortions at boundary points. The second, preferred solution, was to use the Discrete Cosine Transform (DCT) which allows starting and ending values to differ, thereby avoiding the distortions seen in DFT. In addition to not requiring tuning of the padding value, Jockers recommends the DCT approach based upon tests showing DCT curves are in better agreement with human experts.

---

[7] https://cran.r-project.org/web/packages/syuzhet/index.html
[8] http://www.matthewjockers.net/2014/06/05/a-novel-method-for-detecting-plot/
[9] http://www.matthewjockers.net/2015/02/02/syuzhet/
[10] https://annieswafford.wordpress.com/2015/03/02/syuzhet/
[11] http://benschmidt.org/2015/04/03/commodius-vici-of-recirculation-the-real-problem-with-syuzhet/
[12] http://www.matthewjockers.net/2017/01/12/resurrecting/

Swafford also raised the objection of 'ringing artifacts' that are inherent to any type of Fourier Transform modeling of sharp discontinuities with smooth sinusoidals. Sharp impulses and large step function discontinuities cannot be accurately modeled with the low pass filter smoothing technique used in Syuzhet.R. This is a theoretical objection that has not proven to affect the analysis of novels where such extremely odd situations do not arise naturally.[13]

**Objection #2: Inability of simple lexical Sentiment Analysis to accurately account for negation, intensifiers, etc.**

Another criticism raised by Swafford highlights a number of obvious errors inherent to the simple lexical Sentiment Analysis used by Syuzhet.R. These include the inability to correctly assign sentiment scores in cases of negation (not happy), intensifiers (extremely good) and other modifiers like capitalization, punctuation and emojis (SO SAD!!! :o). This is a fundamental limitation of the simple lexical Sentiment Analysis used by Syuzhet. The impact of these errors could prove insignificant by statistical averaging over large enough samples given a relatively unbiased dataset, but we had no data with which to form this conclusion.

Since cases like negation and intensifiers are not uncommon in literary texts, this was the particular technical question we explored and quantified in our paper. If Syuzhet used a fundamentally unsound Sentiment Analysis technique, our literary analysis relying upon it would be rendered inaccurate as well. Prior to our work, we could find no quantitative measure of how much error such statistical simplifications introduce, especially in the case of analyzing text from novels.

We tested the effects of ignoring these intensifiers/modifiers by comparing the simple lexical Sentiment Analysis in Syuzhet against a more sophisticated lexical Sentiment Analysis library called VADER (Valence Aware Dictionary and sEntiment Reasoner).[14] Unlike the naive dictionary approaches used by Syuzhet, the VADER library is augmented with rules to correctly adjust sentiment for the following situations:

- Typical negations (e.g. "not good")
- Use of contractions as negations (e.g. "wasn't very good")
- Conventional use of punctuation to signal increased sentiment intensity (e.g. "Good!!!")
- Conventional use of word-shape to signal emphasis (e.g. using ALL CAPS)
- Using degree modifiers to alter sentiment intensity (e.g. intensity boosters such as "very" and intensity dampeners such as "kind of")
- Understanding many sentiment-laden slang words (e.g. "sux")
- Understanding many sentiment-laden slang words as modifiers such as "uber", "friggin" and "kinda"
- Understanding many sentiment-laden emoticons such as :) and :D
- Translating utf-8 encoded emojis
- Understanding sentiment-laden initialism and acronyms (e.g. "lol")

---

[13] http://www.matthewjockers.net/2015/03/09/is-that-your-syuzhet-ringing/
[14] https://github.com/cjhutto/vaderSentiment

Since Syuzhet was our baseline and already successfully tested by human experts against a number of novels, we couldn't just start using the more sophisticated VADER library. Instead, we needed to keep using the same configuration with Syuzhet, but try to quantify what, if any, are the errors introduced by failing to account for intensifiers, negations, etc. Comparing the results of Sentiment Analysis using VADER and Syuzhet side-by-side on the same novel provides a direct comparison for our case of interest: Woolf's *To the Lighthouse*.

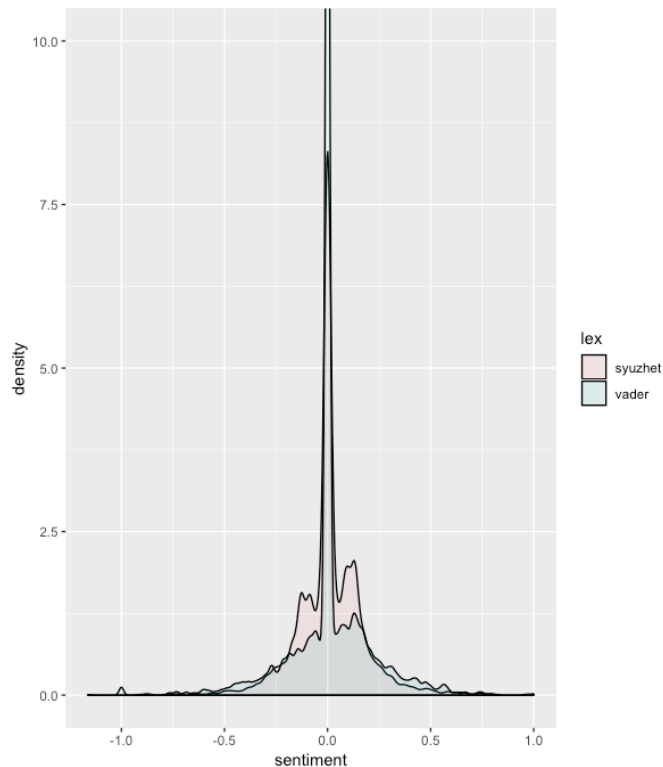

As discussed in the body of this paper, the histogram of per-sentence sentiment scores of our novel using both VADER and Syuzhet above shows an almost identical frequency distribution for both techniques. Both sentiment value distributions share almost identical means, variances and even skews. This suggests that Syuzhet is not statistically distorting sentiment scores by ignoring any of the explicit rules VADER accounts for above. This, in turn, addresses the skepticism Swafford, we, and others have had that simple lexical approaches like Syuzhet with known deficiencies can still produce relatively accurate results on par with more sophisticated methods like VADER.

This result was counter-intuitive, as we expected the more sophisticated VADER library to provide a slightly different interpretation and reveal limitations in the simpler techniques used in Syuzhet. However, simple statistical techniques have a history of doing well and even (at times) outperforming more complex mathematical models (e.g. Deep Neural Networks vs Formal Grammars). Early in his research, when selecting his Sentiment Analysis techniques, Jockers himself noted that he was surprised to find these simple Sentiment Analysis techniques clearly outperformed more advanced Stanford CoreNLP models; indeed, this is why he elected to utilize them. One potential explanation is that more sophisticated models are trained on specialized corpora like product/movie reviews and hand-labeled tweets—a statistically distinct training dataset less appropriate, perhaps, for measuring sentiment in the text of novels.

**Objection #3: Incorrect Sentence Splitting with Dialogue**

Swafford noted that Syuzhet fails to split sentences correctly in passages of dialogue that involve quotations. She provides an example from Dickens *Bleak House* in which three sentences are erroneously parsed as one. Although not a desirable feature, this is the least concerning of her three criticisms. Unless there were distinct patterns of very long passages being erroneously concatenated over key inflection points, we would expect minimal impact. It would be akin to minimally fuzzing an already very stochastic process—that is, a minimal preprocessing smoothing before explicitly smoothing with DCT, LOESS or a sliding window.

To explore the current situation, we first tried to replicate the example Swafford presented. We downloaded a plaintext UTF-8 version of Dickens' *Bleak House* from Gutenberg.org[15] and ran it through the latest version of Syuzhet.R as of Aug 2018. We were unable to reproduce the exact error. Instead only two of the three sentences merged. Furthermore, the sentence splitting seemed grammatically correct due to Dickens frequent stylistic use of double dashes "--" to continue thoughts across independent sentences instead of formally ending them with punctuation marks. Here is our result from trying to replicate the error (note the short second phrase does little to nothing to alter the overall sentiment):

"Mrs. Rachael, I needn't inform you who were acquainted with the late Miss Barbary's affairs, that her means die with her and that this young lady, now her aunt is dead--" "My aunt, sir!""

In Jockers' latest Syuzhet.R release as of Aug 2018, he has code to correct for curly double quotes before calling the library::function textshape::split_sentence[16] to split the novel string into individual sentences. This may have corrected some of the errors Swafford saw with the first release of Syuzhet.R. More importantly, Jockers appears to be relying upon better tested and more widely used standard R NLP packages like textshape to split sentences. This gives us confidence that we are not dealing with non-standard or erroneous parsing.

Nonetheless, to determine if sentence splitting is a statistically significant problem, we examined all sentences longer than 500 characters that were both (a) most likely to exhibit failures to correctly split sentences and (b) more likely to result in a statistical distortion of a critical sentiment inflection point if coincident with such key points. The length of 500 characters as a cutoff was chosen because it was judged as a reasonable compromise between the smallest length of text that could result in distortion of a key inflection point and a reasonable number of sentences that could be manually scored.

Here is our compiled data for both Dickens *Bleak House* and Woolf's *To the Lighthouse*[17] based upon computationally extracting such sentences and manually scoring the error rate. Given that sentences longer than 500 characters are much more likely to exhibit both sentence splitting errors and errors that could statistically dilute key features, we call the percentage of errors found within this subset an 'Upper Bound Percent Error.' That is, there may be more sentence splitting errors below this 500 character threshold, but it is unlikely that they could distort the Sentiment Analysis in a statistically meaningful way given both their short length and our ultimate goal of smoothing the data.

---

[15] http://www.gutenberg.org/cache/epub/1023/pg1023.txt
[16] https://cran.r-project.org/web/packages/textshape/
[17] http://gutenberg.net.au/ebooks01/0100101h.html

|  | Total Sentences | Sentences > 500 characters | Sentence Splitting Errors | Upper Bound Percent Error |
|---|---|---|---|---|
| Dickens | 20,343 | 72 | 1 | 1.4% |
| Woolf | 3,484 | 56 | 1 | 1.78% |

In our samples of sentences longer than 500 characters from these two novels, virtually all were clearly correctly split. In the case of Dickens, it was not so much the use of double quotation marks in dialogue that caused the isolated errors but rather his stylistic use of double dashes "--" to continue ideas across sentences rather than end them with proper punctuation. If this were a problem, it would be trivial to split sentences on double dashes, but we opted not to do this because (a) this appears not to be a statistically meaningful problem and (b) it would result in incomplete and grammatically incorrect sentences likely contrary to the author's original intent.

There was only one example in Dickens that could clearly be argued is an error in sentence splitting. We present it here to demonstrate (a) how syntactically unorthodox a sentence has to be to fool the parser and (b) how little effect such errors have. Below is the example:

"Even," proceeds Sir Leicester, glancing at the circumjacent cousins on sofas and ottomans, "even in many--in fact, in most--of those places in which the government has carried it against a faction--" (Note, by the way, that the Coodleites are always a faction with the Doodleites, and that the Doodleites occupy exactly the same position towards the Coodleites.)  "--Even in them I am shocked, for the credit of Englishmen, to be constrained to inform you that the party has not triumphed without being put to an enormous expense."

Note that if this example were parsed into three separate sentences instead of one, the difference in ultimate sentiment scores would be approximately [-2, 0, -2] instead of [0]. Given that this error is rare, of approximately the same resolution, and results in the same average value after smoothing, we found no need to programmatically correct for such edge cases.

Here is the one sentence splitting error we identified in Woolf's *To the Lighthouse.* Contrary to Swafford's critique, the parser in this case seems to have split more than it should have. We present the original text and the parsed text below.

Original text:

But here, as she turned the page, suddenly her search for the picture of a rake or a mowing-machine was interrupted. The gruff murmur, irregularly broken by the taking out of pipes and the putting in of pipes which had kept on assuring her, though she could not hear what was said (as she sat in the window which opened on the terrace), that the men were happily talking; this sound, which had lasted now half an hour and had taken its place soothingly in the scale of sounds pressing on top of her, such as the tap of balls upon bats, the sharp, sudden bark now and then, "How's that? How's that?" of the children playing cricket, had ceased; so that the monotonous fall of the waves on the beach, which for the most part beat a measured and soothing tattoo to her thoughts and seemed consolingly to repeat over and over again as she sat with the children the words of some old cradle song, murmured by nature, "I am guarding you--I am your support," but at other times suddenly and unexpectedly, especially when her mind raised itself slightly from the task actually in hand, had no such kindly meaning, but like a ghostly roll of drums remorselessly beat the measure of life, made one think of the destruction of the island and its engulfment in the sea, and warned her whose day had slipped past in one quick doing after another that it was all ephemeral as a rainbow--this sound which had been obscured and concealed under the other sounds suddenly thundered hollow in her ears and made her look up with an impulse of terror

Text split into sentences:

[1] "But here, as she turned the page, suddenly her search for the picture of a rake or a mowing-machine was interrupted."
[2] "The gruff murmur, irregularly broken by the taking out of pipes and the putting in of pipes which had kept on assuring her, though she could not hear what was said (as she sat in the window which opened on the terrace), that the men were happily talking; this sound, which had lasted now half an hour and had taken its place soothingly in the scale of sounds pressing on top of her, such as the tap of balls upon bats, the sharp, sudden bark now and then, "How's that?"
[3] "How's that?"
[4] "of the children playing cricket, had ceased; so that the monotonous fall of the waves on the beach, which for the most part beat a measured and soothing tattoo to her thoughts and seemed consolingly to repeat over and over again as she sat with the children the words of some old cradle song, murmured by nature, "I am guarding you--I am your support," but at other times suddenly and unexpectedly, especially when her mind raised itself slightly from the task actually in hand, had no such kindly meaning, but like a ghostly roll of drums remorselessly beat the measure of life, made one think of the destruction of the island and its engulfment in the sea, and warned her whose day had slipped past in one quick doing after another that it was all ephemeral as a rainbow--this sound which had been obscured and concealed under the other sounds suddenly thundered hollow in her ears and made her look up with an impulse of terror."

Sentences [2]-[4] should technically be parsed as one rather than three sentences, yet this was the only error we could find out of the 56 sentences longer than 500 characters in Woolf's *To the Lighthouse*. Contrary to a failure to split sentences with dialogue and quotation marks as Swafford cited in 2015, the current version of Syuzhet seems to err on the side of splitting sentences in such situations. Again, these errors could be corrected with simple rule-based preprocessing, but it makes no sense to do so. These are relatively rare, unbiased and localized errors that have little to no effect on a statistical analysis or the ultimate shape of our smoothed model .